\pdfoutput=1

\documentclass{article}

\usepackage{arxiv}

\usepackage[utf8]{inputenc} 
\usepackage[T1]{fontenc}    
\usepackage{lmodern}        
\usepackage{hyperref}       
\usepackage{url}            
\usepackage{booktabs}       
\usepackage{amsfonts}       
\usepackage{nicefrac}       
\usepackage{microtype}      
\usepackage{graphicx}
\usepackage{authblk}
\usepackage[authoryear]{natbib}

\title{Challenges and Recommendations for Electronic Health Records Data Extraction and Preparation for Dynamic Prediction Modelling in Hospitalized Patients - a Practical Guide: Tutorial}

\author[1]{\small Elena ALBU}
\author[1]{\small Shan GAO}
\author[2]{\small Pieter STIJNEN}
\author[3]{\small Frank E. RADEMAKERS}
\author[4, 5, 6]{\small Bas C T van Bussel}
\author[7, 8]{\small Taya Collyer}
\author[9]{\small Tina Hernandez-Boussard}
\author[1, 4, 10]{\small Laure WYNANTS}
\author[1, 10]{\small Ben VAN CALSTER}

\affil[1]{\footnotesize Department of Development \& Regeneration, KU Leuven, Belgium}
\affil[2]{\footnotesize Management Information Reporting Department, University Hospitals Leuven, Belgium}
\affil[3]{\footnotesize Faculty of Medicine, KU Leuven, Belgium}
\affil[4]{\footnotesize Care and Public Health Research Institute (CAPHRI), Maastricht University, the Netherlands}
\affil[5]{\footnotesize Department of Intensive Care Medicine, Maastricht University Medical Centre+, The Netherlands}
\affil[6]{\footnotesize Cardiovascular Research Institute (CARIM), Maastricht University, The Netherlands}
\affil[7]{\footnotesize Peninsula Clinical School, Central Clinical School, Monash University, Melbourne, VIC, Australia}
\affil[8]{\footnotesize National Centre for Healthy Ageing, Monash University, Melbourne, VIC, Australia}
\affil[9]{\footnotesize Department of Medicine, Stanford University, Stanford, California}
\affil[10]{\footnotesize Leuven Unit for Health Technology Assessment Research (LUHTAR), KU Leuven, Belgium}

\providecommand{\tightlist}{%
  \setlength{\itemsep}{0pt}\setlength{\parskip}{0pt}}

\usepackage{longtable,booktabs,array}
\usepackage{calc} 
\usepackage{etoolbox}
\makeatletter
\patchcmd\longtable{\par}{\if@noskipsec\mbox{}\fi\par}{}{}
\makeatother
\IfFileExists{footnotehyper.sty}{\usepackage{footnotehyper}}{\usepackage{footnote}}
\makesavenoteenv{longtable}

\usepackage{booktabs}
\usepackage{longtable}
\usepackage{array}
\usepackage{multirow}
\usepackage{wrapfig}
\usepackage{float}
\usepackage{colortbl}
\usepackage{pdflscape}
\usepackage{tabu}
\usepackage{threeparttable}
\usepackage{threeparttablex}
\usepackage[normalem]{ulem}
\usepackage{makecell}
\usepackage{xcolor}
\begin{document}
\maketitle

\begin{abstract}
Dynamic predictive modelling using electronic health record (EHR) data has gained significant attention in recent years. The reliability and trustworthiness of such models depend heavily on the quality of the underlying data, which is, in part, determined by the stages preceding the model development: data extraction from EHR systems and data preparation. In this article, we identified over forty challenges encountered during these stages and provide actionable recommendations for addressing them. These challenges are organized into four categories: cohort definition, outcome definition, feature engineering, and data cleaning. This comprehensive list serves as a practical guide for data extraction engineers and researchers, promoting best practices and improving the quality and real-world applicability of dynamic prediction models in clinical settings.
\end{abstract}


\section{Background}\label{background}

Predictive modeling using electronic health record (EHR) data has become increasingly important in enhancing patient outcomes through real-time risk detection and timely interventions. However, the effectiveness of these models is heavily reliant on the quality and structure of the underlying data, which are influenced by the processes of data extraction and preparation. While recent advancements in artificial intelligence and machine learning \citep{vaswani2017attention, lee2019dynamic} have shown promise in this area \citep{tomavsev2021use, tomavsev2019clinically, leeclinical}, significant challenges remain in ensuring that the data used for model development are representative and reliable.

Data collection is part of the routine hospital workflow. Data extraction entails retrieving and structuring raw EHR data from hospital databases using an ETL (extract, transform, load) process, with standards like Observational Medical Outcomes Partnership Common Data Model (OMOP CDM) facilitating structure and terminology consistency. Publicly available ICU and emergency admission datasets \citep{pollard2018eicu, thoral2021sharing, de2023guide, all2019all}, such as MIMIC \citep{johnson2016mimic, johnson2023mimic}, support reproducible research, while hospitals also extract private, not publicly shared data for predictive modeling. Data preparation transforms extracted data into a structured format for predictive modeling, often using R or Python pipelines. Open-source frameworks, whether generalizable \citep{jarrett2023clairvoyance, tang2020democratizing} or specific to MIMIC \citep{gupta2022extensive, wang2020mimic}, aim to establish reproducible workflows for EHR data preparation. Data quality issues, such as inconsistencies and artifacts introduced during extraction and preparation, can compromise the performance of predictive models. Understanding the intricacies of data extraction and preparation challenges is essential for developing robust predictive models that can be reliably integrated into clinical practice.

Structured data quality assessments \citep{lewis2023electronic, miao2023data, thuraisingam2021assessing, nobles2015evaluation} are facilitated by frameworks like Weiskopf and Weng \citep{weiskopf2013methods} and METRIC \citep{schwabe2024metric}. However, researchers often conduct unstructured assessments, documenting ``lessons learned'' from their projects \citep{ferrao2016preprocessing, sauer2022leveraging, arbet2021lessons, maletzky2022lifting, sendak2017barriers, de2024table}. These assessments tend to focus primarily on data cleaning rather than the entire data extraction and preparation process, including cohort definition, outcome definition, and feature engineering. While literature on data quality assessments exists, mitigation strategies remain largely undocumented \citep{honeyford2022challenges}.

We identified and categorized the challenges encountered during EHR data extraction and preparation for predictive modeling. We provided actionable recommendations that can enhance data quality and improve the real-world applicability of dynamic prediction models in clinical settings. By addressing these challenges, we hope to contribute to the development of more reliable and effective predictive modeling frameworks that can ultimately benefit patient care.

\section{Objective}\label{objective}

This article provides a comprehensive list of challenges encountered during data extraction and preparation using electronic health record (EHR) data for developing dynamic prediction models for clinical use. It further proposes recommendations with actionable insights, intended for improving the quality of research and the practical applicability of clinical predictive models. Our insights are drawn from a selective literature review, as well as our experience with various EHR data extractions. This list is intended as a hands-on resource for data extraction engineers (who perform the data extraction) and researchers (who prepare the data for model building) to consult during the extraction and preparation process.

We focus on single-hospital structured data, covering both ICU-specific and hospital-wide extractions. We focus on medium to large-scale extractions, i.e.~generated through structured ETL processes for a large number of extracted items spanning diverse clinical domains (laboratory results, medications, demographics, comorbidities, etc.), and we cover standardized (e.g., OMOP CMD) and non-standard, private extractions. We do not address combined data registries, such as national registries (integrating surveys, general practice, insurance data, and EHR extractions), multi-center data extractions, hospital data collected for clinical trials, or the extraction and processing of unstructured data (e.g., text notes or images). We cover the broader application of dynamic models (or continuous prediction) for offering the most comprehensive framework, although many of the recommendations are applicable to static prediction models (e.g., a single prediction per admission at 24 hours after admission).

Recognizing that EHR software may contain bugs, human errors will occur, and hospital processes will evolve and generally improve over time, we do not provide recommendations for EHR vendors or for modifying hospital workflows, clinical practices, or data recording procedures within EHR systems. When data quality issues, such as those arising from data recording procedures, render the extracted data inadequate for a specific prediction task, we leave it to researchers to assess adequacy, without offering guidance on what constitutes suitable data for a given prediction task.

The section ``EHR Data Flow from Data Collection to Model Building'' explains the typical trajectory of data, for both model building and clinical implementation, setting the stage and providing context for the detailed list of challenges that follows. The section ``Challenges and Recommendations'' lists the challenges and actionable recommendations, which represent the core of the article. These are categorized into four groups: cohort definition, outcome definition, feature engineering, and data cleaning. This is one possible categorization, inspired by stages of data preparation process. The ``Discussion and Conclusion'' section summarizes the recommendations and reflects on their broader implications.

\section{EHR data flow from data collection to model building}\label{ehr-data-flow-from-data-collection-to-model-building}

\begin{figure}
\includegraphics[width=16cm]{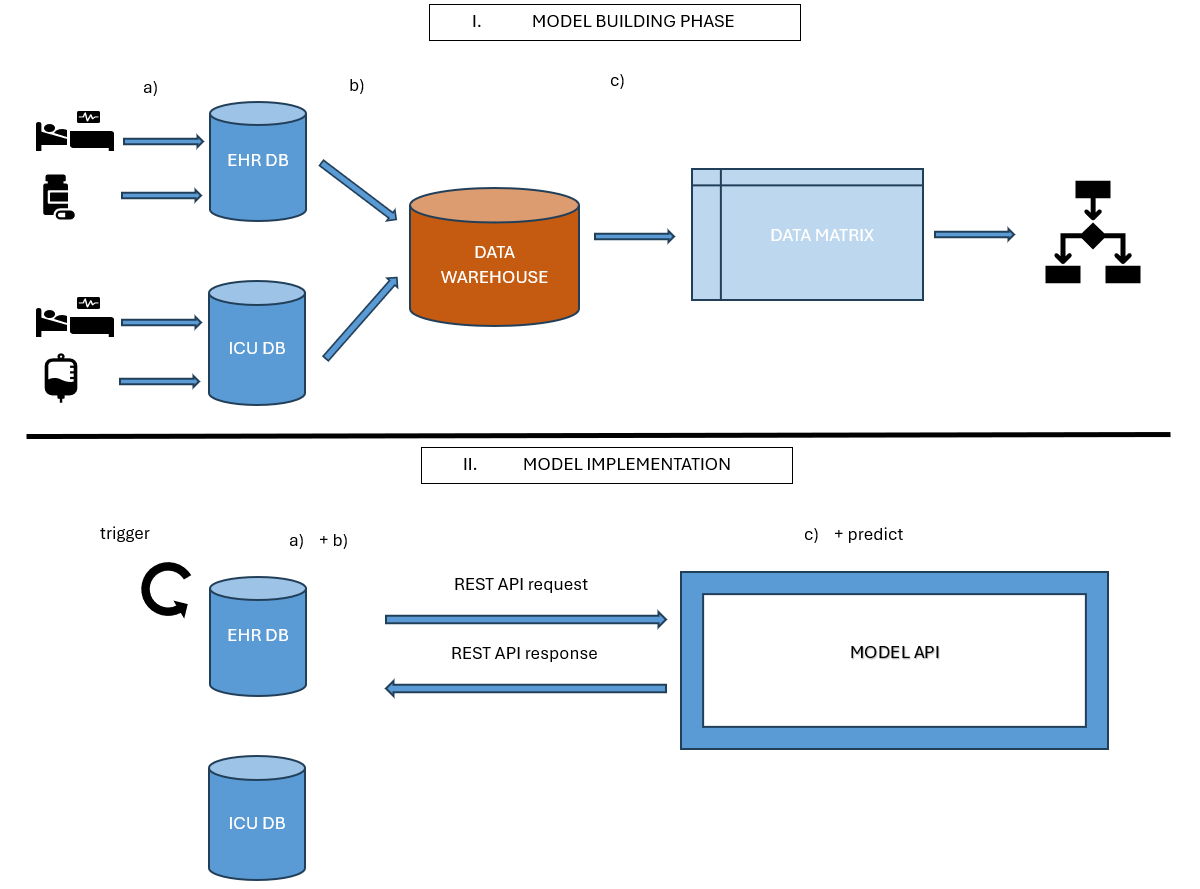} \caption{Data flow for model building pipeline (I.) and model implementation (II.) Two databases are exemplified as data sources, EHR and ICU, although multiple other sources might be used in the hospital's flow and for data extraction. EHR = Electronic Health Record; DB = database; ICU = Intensive Care Unit; REST API = RESTful application programming interface}\label{fig:fig-extract}
\end{figure}

The data flow from the patient's bed to the prediction model building typically follows three stages: data collection (a), data extraction (b) and data preparation (c) (Figure \ref{fig:fig-extract}, I.). Patient data are either manually entered in EHR software modules or collected by devices and stored in one or multiple databases. The data collected serve multiple purposes, e.g., daily bedside clinical work, national benchmarking \citep{van2015data}, or reimbursing the care delivered (a). EHR data are then extracted from relational databases in a simplified (denormalized) format and typically stored in a data warehouse as multiple ``base tables'' per category, each capturing different aspects of patient data and healthcare events. Examples of base tables naming as per OMOP CDM are: PERSON, DRUG\_EXPOSURE, DEVICE\_EXPOSURE, CONDITION\_OCCURRENCE, MEASUREMENT, NOTE, OBSERVATION, but naming and granularity of extraction can vary for non-standard extractions (b). Researchers building a prediction model either have access to the data warehouse or get a copy of all tables or, under specific ethical and legal considerations, a subset of the base tables or a subset of the patients in the data warehouse (e.g., admissions during a specified period of interest for patients undergoing mechanical ventilation). Further, they process the extracted data and bring it in a format on which a prediction model can be built (c).

At model implementation time in clinical practice in the EHR software (Figure \ref{fig:fig-extract}, II.), a trigger (e.g., update of lab results) or a scheduled task (e.g., every 24 hours) will initiate the request for a prediction. Data are already collected (a) for the patient in the EHR database(s). The same logic as for extraction (b) is reproduced (typically by reusing the queries or code used for extraction). Using the data packed in a specific format (e.g., json, xml), the prediction service (typically using a REST API interface for communication) is invoked. Data exchange between the EHR and the prediction service is generally performed using the FHIR (Fast Healthcare Interoperability Resources) standard. The prediction service will further prepare the data (c), invoke the model and return a prediction (and additional information, if foreseen) to the EHR software which will present it to users in the form of alerts or patient flags within the patient's chart. The process is typically logged in the EHR system for monitoring purposes. In summary, at real-time prediction, the data collection (a) happens implicitly and is part of the normal clinical flow; data extraction (b) and data preparation (c) are identically reproduced as for model building.

We provide additional background on some particularities for each of the processes of collection, extraction and preparation, which represent transition phases from one data format to another:

\begin{enumerate}
\def\labelenumi{\alph{enumi})}
\tightlist
\item
  \textbf{Data collection}: The EHR database will not reflect with maximum accuracy the ``true state'' of the patient. First, it will suffer of incompleteness, as not all possible markers and observations can be collected for all patients at all times. The decisions with regards to what data are collected (e.g., which laboratory tests are ordered and performed) is highly dependent on the patient conditions and on the hospital procedures. Data collected during routine clinical practice are generally documented more carefully when also used for national reporting of quality of care indicators. \citep{van2015data} Sufficient data are though collected to support the patient's clinical follow-up and treatment. From this perspective, EHR data differ vastly from data collected for clinical trials, where researchers specify the measurements, measurement methods and collection procedure. Second, nurses and clinicians might have slightly more information than the data collected in the system, either from patient conversations or organizational knowledge. Considering that we do not cover extraction of text notes and reports, tabular data only will always suffer of a level of incompleteness. Nevertheless, tabular data might prove sufficient for specific prediction tasks. Third, both manual data entering and data collected by devices can be at times error prone, software can have bugs and data recording procedures in the system will affect the granularity of observed data and will change over time. These considerations, which can be summarized as data completeness, correctness and currency, as defined by Weiskopf and Weng \citep{weiskopf2013methods}, have to be carefully considered by researchers to assess if EHR data are fit for the prediction goal \citep{thuraisingam2021assessing}.
\end{enumerate}

\begin{enumerate}
\def\labelenumi{\alph{enumi})}
\setcounter{enumi}{1}
\tightlist
\item
  \textbf{Data extraction}: The extracted data might not correctly reflect the EHR database. Although data extraction generally aims for completeness of all clinically relevant data, the extraction process can introduce undesired artifacts or can have its own limitations. Whenever extraction logic bugs are detected, these should be corrected and safeguarded through unit and/or integration tests. Mature extraction platforms, tested through repeated use and proven reliability, will generally be less error prone and researchers can consider the maturity of the extraction platform as a factor influencing the time they will spend on data preparation. The extraction process is typically carried out by a data extraction engineer, data integration developer, data warehouse engineer or ETL developer representing the EHR vendor or the hospital IT department. Hereafter, we refer to this role as the data extraction engineer. They work closely with EHR software developers, hospital IT staff, and clinical personnel to understand database structures and data recording procedures. They perform clinical concept and terminology mapping and document the extracted data.
\end{enumerate}

\begin{enumerate}
\def\labelenumi{\alph{enumi})}
\setcounter{enumi}{2}
\tightlist
\item
  \textbf{Data preparation}: The prepared data might not correctly reflect the extracted data. Undesired artifacts can be introduced by feature engineering or data cleaning. Good documentation of the extraction format and close collaboration between researchers, data engineers and clinical experts will ensure that the data and the features are not misinterpreted. Coding errors can always occur. Time-sensitive data poses an additional challenge to ensure no temporal leaks (using future data to predict past events) are introduced by mistake. Outcome leaks (including outcome information in predictors) and test-train leaks (including information from test patients in the training datasets) can also occur if data are not preprocessed carefully. A good practice is to separate train and test sets first, and apply the data preprocessing separately. A modular organization of the code will facilitate unit testing (testing individual functions of a program in isolation to ensure they work as expected) and easy modifications without introducing new errors. The more complex the data preparation, the more error prone it becomes. The data preparation is performed by a researcher, data scientist, statistician or machine learning engineer, which we will refer to as researcher for the remainder of the paper.
\end{enumerate}

Each stage of processing the data has its own challenges and can introduce new problems, widening the gap between the patient's state and the data used by the prediction model and ultimately impacting the model building (suboptimal model), model evaluation (misleading performance metrics) or model implementation in clinical practice.

\section{Challenges and Recommendations}\label{challenges-and-recommendations}

We list common problems originating in the data collection process (a), the data extraction process (b) and the data preparation process (c). We provide recommendations for mitigation strategies that can be implemented during the data extraction (b) or data preparation (c). We also focus on problems that can impede the identical reproduction of the extraction and preparation at clinical implementation time. We have categorized the challenges and recommendations into four groups: (1) cohort definition (and inclusion/exclusion criteria), (2) outcome definition, (3) feature engineering, and (4) data cleaning, and each group contains problems originating in the collection, extraction or preparation process.

We provide mapping of the items to both the Weiskopf and Weng framework \citep{weiskopf2013methods} and the METRIC framework \citep{schwabe2024metric}, whenever applicable. Weiskopf and Weng do not include dimensions covering the cohort representativeness and completeness of the extracted features, which are important in prediction settings. We will use the Completeness dimension to refer to completeness of data values, as in the original definition \citep{weiskopf2013methods}, as well as completeness of the cohort and of the extracted features (our extension).

While we aimed to make these challenges as generic as possible, we recognize that specific issues are often unique to individual projects and may not apply to all prediction tasks. We advise users to assess the impact of each listed challenge in their specific context. Similarly, the recommendations might not always be universally applicable and depend on the project context. Our guidance remains pragmatic; at times, the best approach may be to ``leave-as-is'' to avoid the risk of overcorrection, which can backfire. While certain corrections may improve alignment with clinical meaning for a specific patient group, they can introduce errors or biases when applied to all patients or be difficult to reproduce at clinical implementation time, affecting the model's performance in clinical use. Following the fitness-for-use principle \citep{miao2023data}, we encourage readers to remain pragmatic and address only the issues relevant to their data and prediction task.

\subsection{Cohort definition}\label{cohort-definition}

Defining the cohort of interest (e.g., hospital-wide population or patients with a specific condition) represents a first step in the prediction task definition (Table \ref{tab:tbl-recommendations-cohort-1}, Table \ref{tab:tbl-recommendations-cohort-2} and Table \ref{tab:tbl-recommendations-cohort-3}). It can also be of interest during the project planning phase. An inaccurately defined cohort can lead to selection bias, resulting in performance estimates that do not accurately reflect the model's performance in clinical practice.

\subsection{Outcome definition}\label{outcome-definition}

Prediction models using EHR data usually focus on in-hospital or post-discharge outcomes, including mortality, length of stay, readmission, acute events (like bacteremia, sepsis, and acute kidney injury), and chronic diseases (such as heart failure, cancer, and cardiovascular disease) \citep{pungitore2023assessment}. We focus on outcomes that can be derived solely from structured EHR data, without linkage to external data sources or extraction of text notes, and are linked to a patient, i.e.~we exclude resource utilization and workflow optimization outcomes (Table \ref{tab:tbl-recommendations-outcome-1} and Table \ref{tab:tbl-recommendations-outcome-2}). Similar to cohort definition, a good practice represents assessing the feasibility of defining the outcome during the project planning phase.

\subsection{Feature engineering}\label{feature-engineering}

Feature engineering is the process of selecting, transforming, and creating features (variables) from the extracted EHR data to ensure that the data are brought into a suitable format for modelling and that relevant information is processed in a meaningful way. It includes data mapping (e.g., mapping medication brand name to generic drug names) and transformation (e.g., grouping medical specialties in meaningful categories), converting timestamped events (e.g., lab results, medication administration, vital signs) into snapshot-based features and data aggregation (summarizing multiple observations or measurements into single features). The feature engineering is typically performed during data preparation, and usually it happens concomitantly with data cleaning. Sometimes feature engineering is also performed at data extraction time, for example, for reducing the data volume for high-frequency time-series data, especially in ICU settings, or for computing scores that are calculated and displayed in the EHR software but not stored in the system. Clinical concept mapping, such as mapping medication to ATC (Anatomical Therapeutic Chemical) codes, laboratory tests to LOINC (Logical Observation Identifiers, Names, and Codes) codes or procedures and clinical observations to SNOMED CT (Systematized Nomenclature of Medicine Clinical Terms) can occur at different stages, including within the EHR software, during data extraction (e.g., OMOP CMD Standardized Vocabularies), or as part of data preparation.

We distinguish between generic feature engineering (Table \ref{tab:tbl-recommendations-feat-not-ts-1} and Table \ref{tab:tbl-recommendations-feat-not-ts-2}), which deals with the availability and mapping of clinical items, and time-sensitive feature engineering (Table \ref{tab:tbl-recommendations-feat-ts-1} and Table \ref{tab:tbl-recommendations-feat-ts-2}), which deals with data aggregation for which correct processing of timestamps is critical and that can result in temporal leaks (i.e.~data available at a specific time for training the model, but not available in the system at that timestamp). Time-sensitive feature engineering is critical for dynamic prediction models, can be useful also for some static models (e.g., predictions at 24 hours after admission), and has less impact on models with prediction trigger at the end of the admission (e.g., readmission prediction). We acknowledge that some minimal temporal leaks might not be very detrimental for the model performance or for its applicability, while others can have a large impact. However, following the ``do it right the first time'' principle, good understanding of the extracted timestamps and correct handling of date/times during data extraction and preparation can safeguard against future problems, big or small.

As for the previous two groups, certain aspects of feature engineering can be evaluated during project planning, at least for the key features relevant to the prediction task and, at a minimum, for their availability. For instance, if a key feature was only recorded for a limited period before being discontinued or if timestamps of key features are unavailable for a dynamic prediction task, the project may become jeopardized.

\subsection{Data cleaning}\label{data-cleaning}

The data cleaning process constitutes of identifying and correcting data issues that could negatively impact the performance or applicability of prediction models. Errors in EHR data can arise from manual entry mistakes, improperly connected or malfunctioning devices, or bugs within the EHR software. Such errors are typically uncovered during data exploration and addressed during data preparation. Tools like the Data Quality Dashboard \citep{blacketer2021increasing} have been developed to support and streamline the data cleaning process. Artifacts can also be introduced during data extraction or during data preparation. Unit tests for both the ETL process and the data preparation pipeline can safeguard against introducing additional errors.

Overzealous correction of errors, such as manual correction during data preparation of every error encountered might not necessarily prove beneficial for prediction tasks, when such corrections cannot be programmatically reproduced during model implementation in clinical settings. This discrepancy can lead to a situation in which training data accurately represents the true patient state, while real-time predictions rely on erroneous data, resulting in reduced predictive performance. Although measures can be taken to address common and documented errors from previous studies or identified during data exploration (Table \ref{tab:tbl-recommendations-clean-1} and Table \ref{tab:tbl-recommendations-clean-2}), it is impossible to anticipate and guard against all future errors. For example, new EHR software versions and updates may introduce new bugs, even as older bugs are resolved.

\begingroup\fontsize{7}{9}\selectfont

\begin{longtable}[t]{>{\raggedright\arraybackslash}p{3cm}>{\raggedright\arraybackslash}p{14cm}}
\caption{\label{tab:tbl-recommendations-cohort-1}COHORT Recommendations Related to Identifying Admissions. ICU: Intensive Care Unit; EHR: Electronic Health Record; HIS: Hospital Information System; METRIC = METRIC cluster and dimension mapping; WW = Weiskopf and Weng dimension mapping}\\
\toprule
Problem & Description and recommendation\\
\midrule
Linking identifiers & DESCRIPTION: As data sources are relational databases, the extraction process deals with a large number of identifiers. Patient, admission, physician and even bed \citep{maletzky2022lifting} identifiers are used for linking clinical items. The process of anonymizing identifiers can also introduce errors. Furthermore, identifiers can differ in different systems (ICU vs. non-ICU). The process of denormalization at data extraction obscures a large number of identifiers. Nevertheless, generally patient and admission identifiers (anonymized) remain part of the extracted data. Incorrect linking of identifiers can result in incorrect extracted cohort. \newline  RECOMMENDATION: Correct linking of obscured identifiers can be tested during data extraction, for example by verifying date of birth of two linked patient items from two systems match \citep{ferrao2016preprocessing}. \newline  METRIC: Representativeness - depth of data - coverage; WW: Completeness\\
Cross-linking patients and admissions across various data sources creates inconsistencies & DESCRIPTION: Patient demographics and admission data are typically stored in a HIS (Hospital Information System), which acts as a 'master' system for patient data and shares these data with other specialized clinical software modules, such as the emergency system, ICU system, and EHR modules used in other wards. In a flawless integration between systems, all specialized clinical modules maintain identical copies of patient and admission information as the 'master' HIS system. However, discrepancies might exist, especially in historical data. Issues such as failed patient merges, or demographic updates that do not correctly propagate across all systems, can lead to inconsistencies. Additionally, data migration from previous software vendors may introduce discrepancies. For example, a cohort including all patients in the emergency department over 60 might be differently identified based on HIS patient demographics than on the demographics from the emergency software. While these situations should normally be rare, they may be more common in historical data. \newline  RECOMMENDATION: If large discrepancies have been created due to historical reasons, these would ideally be addressed during data extraction, considering the knowledge on historical system integration problems. If discrepancies persist in current clinical practice at the time of model deployment for clinical use, cohort identification and patient demographics may need to reflect the system in which the prediction model is used, e.g., if the model is deployed in the emergency department, patient data may need to reflect the information available to emergency physicians. \newline  METRIC: Representativeness - depth of data - coverage; WW: Completeness\\
Invalid admissions, not reflecting real patients, are present in extraction & DESCRIPTION: Admissions that do not represent real patients are present in EHR databases \citep{de2024table}. At minimum, 'test patients' on which new EHR software versions are tested will be present in most EHR databases. Other examples can include: 'ghost' admissions resulting from incorrect data migration from previous EHR vendor database, or administrative admissions for beds or equipment to enable lab sample collection. Such invalid admissions might or might not be linked to a patient record. Such admission can occur only during training data timeframe (for historical reasons, such as data migration problems), or both at training and clinical implementation time (for test patients).\newline  RECOMMENDATION: Test patients can be easily filtered during data extraction by excluding a list of known test patient or admission identifiers; these can be difficult to detect during data preparation. Detecting invalid admissions due to historical problems requires knowledge of the database and the historical hospital workflows; generally, these are best handled during data extraction. If invalid admissions are detected during data exploration, the root cause of the underlying problem should better be discussed with data extraction engineers. Inclusion of such admissions at clinical implementation would generally not represent a real danger, as healthcare professionals are aware that predictions created for admissions that do not represent a real patient can be safely disregarded, and in most workflows, these would not interrupt their workflow. The impact on clinical implementation should though be evaluated case by case. \newline  METRIC: Informativeness - redundancy - uniqueness; WW: NA\\
Duplicated admissions present in extraction & DESCRIPTION: The same patient admission is extracted twice, with different identifiers. This can occur due to extraction in separate batches or identifiers anonymization and leads to a biased cohort definition. Duplicated admissions can also be recorded in the system for administrative or funding purposes. \newline  RECOMMENDATION: Such problems can be captured in the data extraction framework using unit or integration tests. At data preparation, tables can be checked for identical duplicated records. \newline  METRIC: Informativeness - redundancy - uniqueness; WW: NA\\
Restricting extraction to hospitalized patients may lead to omission of useful patient data & DESCRIPTION: Often, data extraction is restricted to hospitalized patients (also called inpatient admissions), excluding outpatient (consultations) or day-care admissions. Some hospitalized patients may have relevant data recorded during previous outpatient admissions. Depending on hospital workflows, certain prediction tasks, such as those for planned surgeries, may benefit from pre-admission data recorded during outpatient visits. For instance, lab samples ordered one or two days before an inpatient admission may be linked to an outpatient visit and could be clinically relevant for the prediction task. \newline  RECOMMENDATION: Extraction can be extended to all hospital admissions (inpatient, outpatient and day-care) and researchers can decide if outpatient or day-care admissions contain useful data for the prediction task. \newline  METRIC: Representativeness - depth of data - coverage; WW: Completeness\\
\bottomrule
\end{longtable}
\endgroup{}

\begingroup\fontsize{7}{9}\selectfont

\begin{longtable}[t]{>{\raggedright\arraybackslash}p{3cm}>{\raggedright\arraybackslash}p{14cm}}
\caption{\label{tab:tbl-recommendations-cohort-2}COHORT Recommendations Related to Inclusion/Exclusion Criteria. EHR: Electronic Health Record; ICD: International Classification of Diseases; METRIC = METRIC cluster and dimension mapping; WW = Weiskopf and Weng dimension mapping}\\
\toprule
Problem & Description and recommendation\\
\midrule
Incorrect or incomplete mappings leading to incorrect cohort definition applied at extraction & DESCRIPTION: For some prediction tasks only a subset of hospitalized patients is of interest, e.g., patients with a central catheter / urinary catheter, patients undergoing surgery (for postoperative outcomes), patients on mechanical ventilation, or patients with a specific diagnose or condition. The items that identify a device (a mechanical ventilator, a catheter) or the definitions used for defining a patient condition might differ between medical wards. Correct identification of these items during data extraction might be challenging and inadvertently items may be missed, creating either gaps in data and incorrectly excluding admissions. Often it is difficult to spot such incorrect exclusions during data preparation, as the researcher does not have access nor knowledge of the EHR system configuration, hospital workflow and data recording procedures in each department. Incorrect exclusion can be due to many factors, such as typos (an item has been configured with a typo for some period in the system, e.g., centraal vs. central venous catheter) or different recording procedures in specialized wards (e.g., dialysis catheters are recorded in a special form in the Nephrology department), ultimately leading to sample selection bias \citep{sauer2022leveraging}. For identifying patient conditions, multiple sources can be used, such as ICD codes or discharge diagnoses. \newline  RECOMMENDATION: If ethical approval can be granted to extract all admissions within the timeframe of interest, without applying other specific exclusion criteria, cohort definition can be done during data preparation, as a collaborative effort between data extraction engineers, researchers and clinical representatives. This allows researchers to perform additional verification, improving the cohort definition through multiple reviews. Whenever this is not possible, the data extraction engineers would have to work closely with clinical representatives for correct cohort definition. Correct clinical concept and terminology mapping becomes crucial. \newline  METRIC: Representativeness - depth of data - coverage; WW: Completeness\\
Inclusion/exclusion criteria require patient history prior to the study period & DESCRIPTION: Inclusion/exclusion criteria based on a history of chronic conditions, such as osteoarthritis \citep{thuraisingam2021assessing}, may not be fully recorded in the EHR system or might require extracting data from a period preceding the study timeframe, for example, a 'look-back period' of 3 to 5 years, as historical conditions are associated with previous admissions. \newline  RECOMMENDATION: Assessing the completeness of data used to define the inclusion/exclusion criteria prior to study initiation can be done through external source linking to gold standard data sources \citep{thuraisingam2021assessing}. Alternatively, if historical diagnoses relevant to current admission are available in (structured or unstructured) clinical notes recorded shortly after admission, these can be used. \newline  METRIC: Representativeness - depth of data - coverage; WW: Completeness\\
Restrictive inclusion/exclusion criteria applied at extraction limit the scope of missing data imputation & DESCRIPTION: Imputation of missing values (e.g., lab results) might benefit from the extraction of a larger cohort than the one strictly of interest. Similarly, if the admissions are further split in episodes of interest for prediction, which are segments of time during a patient admissions when a patient is considered at risk for the outcome of interest (e.g., catheter episodes, mechanical ventilator episodes), the imputation can be done in the largest possible context (using all patient-days during an admission instead of only the episode-patient-days). Performing missing data imputation in a larger context, would in theory, be beneficial, especially if the cohort of interest is rather small, although there is no evidence of this topic being studied. \newline  RECOMMENDATION: If ethical approval can be granted to extract all hospitalizations during a specified time frame, cohort definition using inclusion/exclusion criteria can be done during data preparation, eventually after missing data imputation. \newline  METRIC: Representativeness - depth of data - coverage; WW: Completeness\\
\bottomrule
\end{longtable}
\endgroup{}

\begingroup\fontsize{7}{9}\selectfont

\begin{longtable}[t]{>{\raggedright\arraybackslash}p{3cm}>{\raggedright\arraybackslash}p{14cm}}
\caption{\label{tab:tbl-recommendations-cohort-3}COHORT Recommendations Related to Episodes for Interest for Prediction. ICU: Intensive Care Unit, EHR: Electronic Health Record, ICD: International Classification of Diseases, CPT: Current Procedural Terminology, PEEP: Positive End-Expiratory Pressure, FiO2: Fraction of Inspired Oxygen; METRIC = METRIC cluster and dimension mapping; WW = Weiskopf and Weng dimension mapping}\\
\toprule
Problem & Description and recommendation\\
\midrule
Discontinuity or fragmentation of the patient stay due to cross-linking across different data sources & DESCRIPTION: Whenever the cohort definition is done at extraction, e.g., admissions for patients with a specific device, and data are extracted from multiple databases (ICU and hospital-wide EHR), data for the entire hospital admission has to be extracted from both systems. For example, a patient admitted through emergency, transferred to ICU and then to the Nephrology ward has a dialysis catheter only during the ICU stay. Extracting only the data during the ICU stay might not be sufficient for the prediction task. Data prior to ICU contains predictive information. Often, for devices like catheters, the patient is considered at risk of catheter associated infections 48 hours post catheter removal, therefore the patient is still at risk for the event of interest for prediction post-ICU stay, during the stay at the Nephrology ward. Alternatively, discontinuities can happen for extracted data when patient is transferred between wards due to overlooked mapping of clinical items from more systems (e.g., virtually a patient has a central catheter in the general ward, is transferred to ICU without the presence of the  central catheter, and gets transferred back to the general ward, with re-appearance of the central catheter). \newline  RECOMMENDATION: Whenever the hospital admission data is located in multiple data sources (ICU and EHR), data relating the entire hospital admission should be extracted, even if the inclusion criteria are met only in one of the data sources. \newline  METRIC: Representativeness - depth of data - coverage; WW: Completeness\\
Temporal leaks in cohort definition & DESCRIPTION: A subgroup of patients with a specific condition might be of interest for the prediction task, e.g., cardiac patients, but predictions are made during the entire hospital admission. Defining the subgroup using criteria that are not available at the start of admission (or at the first prediction trigger), such as discharge ICD codes (coded typically after patient discharge), or patient presence in the cardiology ward (while the patient was admitted through emergency, was transferred to the ICU and then to the cardiology ward), will affect the applicability of the model. \newline  RECOMMENDATION: The cohort should be defined using data available at first prediction trigger, e.g., data present at admission (e.g., admission reason) or data available during previous admissions (e.g., patients with ICD codes indicative of cardiovascular conditions during previous admissions). Alternatively, the first prediction trigger can be defined when data becomes available: predictions start at patient transfer to the cardiology department and end at patient discharge. Inclusion/exclusion criteria should never be defined using data after the prediction trigger, e.g., excluding patients who died for predictions during the patient's hospitalization. \newline  METRIC: Timeliness - currency; WW: Currency\\
Definition of episodes of interest & DESCRIPTION: The patient follow-up time relevant for the prediction task might be different from the admission timeframe, from admission time until discharge time. This can either mean fragmenting hospital admissions in episodes of interest: catheter episode, mechanical ventilation episode, postoperative follow-up, or, conversely, merging subsequent overlapping hospital admissions \citep{miao2023data, de2024table}. Defining the episodes of interest at data extraction time by using date filtering in large queries can prove error-prone. \newline  RECOMMENDATION: Defining the episodes of interest during data preparation, rather than during data extraction, might prove less error-prone and comes with the advantage of benefiting from a larger patient context for prediction. Extraction of full admissions is recommended. \newline  METRIC: Representativeness - depth of data - coverage; WW: Completeness\\
Episode fragmentation & DESCRIPTION: Definition of episodes of interest, such as catheter episode and mechanical ventilator episodes can be done by start and end times: catheter placement and catheter removal, start and end time of mechanical ventilator, whenever these are available in the system and recorded at their time of occurrence. We warn that using the coded procedures (CPT - Current Procedural Terminology - codes) start and end times might hinder the model's applicability, as these are generally used for billing and registered in the system post-factum, sometimes at the end of the admission. Whenever start and end times are not available for prediction time, the registration of specific observations (e.g., catheter monitoring observations) or parameters (PEEP and FiO2 for mechanical ventilation) identify an episode. These observations or parameters can exhibit gaps in registration. Observations can be recorded manually in the EHR system (catheters), or recorded by devices and integrated in the EHR system with a prespecified granularity, like hourly or every minute (mechanical ventilation). Whenever recorded manually, they can display weekend effects. i.e. less or no values recorded during weekends due to understaffing and leading to data gaps. Whenever recorded by devices, the fragmentation might depend on the granularity of measurements integrated, which can be set differently per monitoring device. \newline  RECOMMENDATION: Exploration of the episodes can be done during data preparation, by checking the number of episodes per patient, the episodes length and time between episodes. Comparison to procedure coded placement and removal time can be done, whenever these are available, but we discourage using procedure codes if these are not registered in the system in real time. \newline  METRIC: Measurement process - completeness; WW: Completeness\\
\bottomrule
\end{longtable}
\endgroup{}

\begingroup\fontsize{7}{9}\selectfont

\begin{longtable}[t]{>{\raggedright\arraybackslash}p{3cm}>{\raggedright\arraybackslash}p{14cm}}
\caption{\label{tab:tbl-recommendations-outcome-1}OUTCOME Recommendations for Outcomes Directly Available in the Extraction. ICD: International Classification of Diseases; COVID-19: Coronavirus Disease 2019; CAM: Confusion Assessment Method; DRS: Delirium Rating Scale; AKI: Acute Kidney Injury; NA: Not Applicable; METRIC = METRIC cluster and dimension mapping; WW = Weiskopf and Weng dimension mapping}\\
\toprule
Problem & Description and recommendation\\
\midrule
Outcomes derived based on ICD codes are not timestamped and often unreliable (under- or upcoded). & DESCRIPTION: Diagnoses are typically recorded by coders using ICD-9 or ICD-10 coding system for reimbursing purposes, typically after patient discharge, without reflecting the time when the diagnose was established. In some cases through, these could be coded during the patient's admission, although these situations are rare. Coded diagnoses are subject to undercoding, upcoding, or creep (misspecification or miscoding) \citep{o2005measuring} or can represent the diagnostic differential (other possible conditions considered before confirming a final diagnosis) rather than the actual diagnosis \citep{arbet2021lessons, ghassemi2020review}. Coding practices can vary based on admission types; for example,  ICD codes for emergency patients may be restricted only to the presenting complaint, while ICD codes for inpatients may reflect chronic diseases and comorbidities which are unrelated to the presenting complaint. Typically, there will be more than one code that clinically identifies a disease (e.g., chronic kidney disease) and these have to be carefully selected using clinical knowledge and mapped to the outcome label. Hospitals have also migrated from ICD9 to ICD10 historically, which might have introduced a period of discontinuity in the availability the diagnosis codes, and the coding practices might change after the transition to ICD-10 \citep{o2005measuring}. ICD codes might also evolve over time, e.g., introduction of a code for COVID-19 in 2020. Due to lack of timestamps, ICD codes are generally inappropriate for prediction of acute events during a patient's admission. Differences in coding practices can introduce bias in the outcome defined by the ICD codes. \newline  RECOMMENDATION: When using ICD codes to define the outcome during a hospital admission (e.g., acute events), it is important to understand the hospital's coding practices, including potential under or upcoding and ensure that relevant timestamps for the event can be established. Additionally, it is important to explore temporal trends, as the coding practice might have changed over time. \newline  METRIC: Measurement process - source credibility - traceability; Timeliness - currency; WW: Currency\\
Outcomes retrospectively assessed by clinical experts are labour-intensive and subject to human error. & DESCRIPTION: Clinical experts can determine outcomes by reviewing patient records retrospectively, either systematically at regular intervals (e.g., an infection preventionist reviewing records monthly to report hospital-acquired infections to national public health institutes) or specifically for a prediction task (e.g., labeling patient records in extracted data). Both approaches require significant time and effort and are prone to errors. \newline  RECOMMENDATION: One way to reduce human error is to involve a group of experts to reach a consensus on outcome assessment. However, this makes the task even more labour-intensive. Additionally, research or hospital teams may want to evaluate the model’s performance after clinical implementation, for example one year post-deployment. In this case, outcomes will need to be assessed by the same experts using the same procedure as during model training, which is rarely feasible. \newline  METRIC: Measurement process - human-introduced error - noisy labels; Measurement process - completeness; WW: Correctness\\
Outcomes documented in the system are subject to mislabeling and delay in registration & DESCRIPTION: Assessments done during the patient's stay, e.g. delirium assessment by a neurologist using Confusion Assessment Method (CAM) or Delirium Rating Scale (DRS), can be used to define the outcome. These can though be subject of different documentation practices in different wards. For example, delirium might be consistently evaluated on the neurology ward and after surgery, but inconsistently in other wards or in patients that did not undergo surgery. Unless meticulously documented, clinical assessments are subject to inconsistencies in different patient groups or across medical wards. There may also be delays in documentation compared to the clinical assessment time (e.g., documented at the end of the day), which can impact time-sensitive outcomes. \newline  RECOMMENDATION: Understanding the assessment practices of the event(s) identifying the outcome is essential in defining its reliability. Understanding the documentation practices (and the registration delay compared to the actual event time) can also be important, depending on how time-sensitive the predictions are, how often the predictions are renewed and the prediction time horizon. This information is typically retrieved from clinical experts, as documenting every single clinical assessment practice during data extraction would be overwhelming. Manual relabelling by clinical experts post extraction, based on patient assessments, possibly combined with other data such as clinical notes, can prove as a viable solution for some outcomes, although it remains a labour intensive task. \newline  METRIC: Informativeness - understandability; Measurement process - human-introduced error - noisy labels; Measurement process - completeness; Timeliness - currency; WW: Correctness; Currency\\
Specific prediction tasks require the extraction of auxiliary outcomes and/or competing events & DESCRIPTION: Deep neural network models may benefit of the use of auxiliary outcomes (multivariate outcomes, e.g., predicting the levels of creatinine, urea nitrogen, sodium, potassium, chloride, calcium and phosphate, along with AKI) as a form of regularization \citep{tomavsev2021use, tomavsev2019clinically}. Including competing events (like death, discharge, device removal) are sometimes modelled when using regression models \citep{gao2024comparison, thuraisingam2021assessing}. \newline  RECOMMENDATION: Whenever applicable for the prediction task, the availability of auxiliary outcomes and/or competing event at extraction should be checked. \newline  METRIC: Informativeness - feature importance; WW: Completeness\\
Label leakage due to temporal leakage & DESCRIPTION: Whenever the prediction time is too close to the event time, strong features will rather detect than predict the outcome which is already known to the healthcare providers. Such an example is mortality prediction at the disconnection of the mechanical ventilator as a family decision to withdraw care at a terminal stage of the ilness \citep{ghassemi2020review}. Another example is including timepoints after the occurence of the event of interest, e.g., sepsis, in model training and evaluation. \citep{wong2021external} \newline  RECOMMENDATION: Careful definition in the outcome label can alleviate such situations. For example, contact with palliative care, transfer to palliative care or death can be used as outcome definition for mortality prediction. The data preparation process should ensure post-event data is excluded when necessary. \newline  METRIC: NA; WW: NA\\
\bottomrule
\end{longtable}
\endgroup{}

\begingroup\fontsize{7}{9}\selectfont

\begin{longtable}[t]{>{\raggedright\arraybackslash}p{3cm}>{\raggedright\arraybackslash}p{14cm}}
\caption{\label{tab:tbl-recommendations-outcome-2}OUTCOME Recommendations for Outcomes Derived Using Clinical or Surveillance definitions. ICD: International Classification of Diseases; AKI: Acute Kidney Injury; CLABSI: Central Line-Associated Bloodstream Infection; VAE: Ventilator-Associated Event; CDC: Centers for Disease Control and Prevention; NA: Not Applicable; METRIC = METRIC cluster and dimension mapping; WW = Weiskopf and Weng dimension mapping}\\
\toprule
Problem & Description and recommendation\\
\midrule
Outcomes derived using clinical or surveillance definitions often require extraction of multiple data sources & DESCRIPTION: Clinical or surveillance definitions provide a more robust alternative compared to ICD codes or documentation of clinical assessments for certain acute events (e.g., AKI, CLABSI, VAE). It comes with advantages: the event occurrence/onset timestamp can be established and the same definition is used over the entire training and test data, but its consistency might still depend on the recording practices of the items used in the definition. Furthermore, these items span across multiple data sources, e.g., for CLABSI, admission data, patient age data (in months, with focus in neonates), laboratory, microbiology, vital signs and imaging reports are necessary to fully comply to the definition. \newline  RECOMMENDATION: All data sources and their completeness should be checked for deciding the appropriateness of the extracted data for outcome derivation. This is typically done during data preparation. Additionally, we recommend extracting both the clinically relevant timestamp as well as the timestamp when data are recorded in the system, whenever these differ (examples in Table 4), and using the clinically relevant timestamp for outcome derivation, e.g. the sample collection time for creatinine when calculating AKI outcome. \newline  METRIC: Measurement process - completeness; Representativeness - depth of data - granularity; WW: Completeness\\
Clinical or surveillance definitions are complex and their implementation is prone to variations & DESCRIPTION: The definition logic for clinical or surveillance definitions are documented in large documents, typically issued by health authorities like CDC, and the logic is generally complex. As these are replicated in code during data preparation using their documentation as the coding requirements, good understanding of the definition is necessary. For example, the VAE definition is time-sensitive and there are multiple corner cases (parts of the definition implementation that require a specific adapted logic). The definitions can be subject of interpretation and there will be variation if different people implement it based on the same documentation. Moreover, coding bugs can occur when implementing complex logic. \newline  RECOMMENDATION: Unit tests can be foreseen in data preparation for corner cases and time-sensitive interpretation. Good understanding of the definition and its clinical interpretation can be achieved by communication with clinical experts. Code reviews can also capture misinterpretations. Whenever other definitions are historically available (e.g., ICD codes, CLABSI definition manually assessed and documented by infection preventionists), these can be used for detecting deviations. Some detected deviations might help with correcting programming errors, but not all deviations will necessarily constitute a problem. \newline  METRIC: NA; WW: NA\\
Outcomes derived using clinical or surveillance definition can be affected by missing or erroneous data & DESCRIPTION: The retrospective calculation of clinical or surveillance definition relies on the availability of different items in the system, which in turn relies on the hospital process of recording these items. For example, CLABSI or VAE events include fever as a symptom check, which in turn, can be derived by using temperature values. Temperature values can be missing and errors in recording these values will be present. \newline  RECOMMENDATION: The data recording frequency and its impact on the outcome calculation should be assessed. The necessity of cleaning or imputing the data prior to outcome calculation should be assessed. Alternatively, assumptions can be made (e.g., microbiology results are always ordered in the presence of fever) to simplify the definition, but such assumptions should be carefully reviewed with clinical experts. \newline  METRIC: Measurement process - completeness; Representativeness - depth of data - granularity; WW: Completeness\\
\bottomrule
\end{longtable}
\endgroup{}

\begingroup\fontsize{7}{9}\selectfont

\begin{longtable}[t]{>{\raggedright\arraybackslash}p{3cm}>{\raggedright\arraybackslash}p{14cm}}
\caption{\label{tab:tbl-recommendations-feat-not-ts-1}FEATURE ENGINEERING Recommendations (not time-sensitive) Related to Clinical Concept Mapping. AVPU = Alert, Voice, Pain, Unresponsive; EHR: Electronic Health Record; ICU: Intensive Care Unit; OMOP CDM: Observational Medical Outcomes Partnership Common Data Model; ICD: International Classification of Diseases; ETL: Extract, Transform, Load; METRIC = METRIC cluster and dimension mapping; WW = Weiskopf and Weng dimension mapping}\\
\toprule
Problem & Description and recommendation\\
\midrule
Inconsistencies due to data extraction of the same clinical items from various databases & DESCRIPTION: There will be high variation in data extraction procedure when data sources differ. A typical example are extractions from EHR (hospital-wide) and ICU system of vital signs, devices (e.g., urinary catheter, dialysis catheter), scales and scores, signs and symptoms or other care observations. These often follow different structures (e.g., units, such as cm for neonates length or m for adults height, are available or units are implicitly considered the same unit; devices registered with start and end time in ICU, but only through a 'device present' observation in the EHR), require different mappings, have different significance (e.g., different pain scales used in different wards or age groups) or can be recorded only in one system (e.g., specific scales are mostly used in ICU, like Glasgow Coma Scale), or recorded in a structured way in one system and in free text in another system (e.g., 'patient is alert' in a free text note vs. AVPU  scale). \newline  RECOMMENDATION: One approach is to extract the same items in a harmonized format. Another approach is to extract separate base tables for each data source. The maximum possible information should be extracted from all systems, regardless if data are harmonized during extraction or during data preparation. (e.g., extract units or start/end date and time if available in one system only). If data sources are harmonized at extraction, include the data source, as cleaning steps might differ for different data sources (e.g., cleaning temperatures recorded by devices in ICU can be different than cleaning errors in manually entered temperatures in other wards). Standardization at extraction using OMOP CMD helps in harmonizing both the data structure and the clinical concepts terminology across different data sources. \newline  Consistency - rule-based consistency - syntactic consistency; WW: NA\\
Same clinical concept available in different tables in the same database & DESCRIPTION: Differences in registration can occur even when data originate in the same database. The registration of specific clinical concepts can deviate from the standard procedure in specialized wards. For example, catheter information may be recorded in a standard structured way for all wards (using nursing observations), but dialysis catheters are recorded in a specialized software module in nephrology department, and implicitly a separate database table. The same can happen with specialized software modules deployed over time, and items being recorded in a database table in older data and in another table in newer data. Additionally, some fields might represent patient reported data while others represent clinically assessed data \citep{arbet2021lessons} and depending on the research question, either both or only the clinically assessed data can be relevant. \newline  RECOMMENDATION: During the clinical concept mapping phase (typically done at data extraction) these recording practices and historical transitions have to be captured accurately from each underlying table. Data exploration during data preparation should reveal only changes in hospital processes if the mapping has been correctly done during extraction. \newline  Consistency - rule-based consistency - syntactic consistency; WW: NA\\
Clinical concept mapping differs between wards or over time & DESCRIPTION: Even when the same clinical concept is recorded in the same table, configuration catalogues might have changed over time or might differ in different wards. Configuration changes (e.g., CVC vs. central venous catheter) or even small typos (e.g., centraal vs. central venous catheter) can create gaps in data if not mapped correctly. \newline  RECOMMENDATION: The clinical concept mapping requires knowledge of the historical use of the EHR software in the hospital, including changes in configuration catalogues, and is best achieved through collaboration with IT specialists or clinical experts. Typically, this is done at extraction, as a step in the ETL process. OMOP CDM provides standardized mapping, but does not solve the problem of correctly detecting all original items to be mapped. \newline  Measurement process - human-introduced error - carelessness; Consistency - rule-based consistency - syntactic consistency; WW: Correctness\\
Partially overlapping data regarding the same clinical concept available in multiple extracted tables & DESCRIPTION: Often, multiple extracted tables contain the same clinical information. For example, diagnoses might be coded using ICD codes at the end of the admission, but they might also be available from care pathways recorded during the admission. Alternatively, ICD-9 and ICD-10 coding might coexist \citep{miao2023data}. Medical procedures are coded using CPT codes, typically after discharge, but information on the performed procedures might be available in structured fields in EHR during the patient admission. Medication orders, prescription and administration might represent three different extracted tables. \newline  RECOMMENDATION: Provided clear extraction documentation, the decision either merging extracted items in one feature or making a choice of a source against the other is generally taken during data preparation. Data source agreement exploration can be done during data preparation. \newline  Informativeness - redundancy - conciseness; Consistency - rule-based consistency - compliance; Documentation; WW: Concordance\\
Changes in hospital processes over time result in gaps in the extracted features & DESCRIPTION: Gaps in recording features over large periods of time (months or years) can occur. The EHR software or specific software modules are generally deployed ward by ward, and such gaps in data can be seen in the first years of the extracted period, before hospital-wide adoption of the software (e.g., if emergency department only started using the software since 2014 but data extraction starts in 2012). This, of course, depends on when processes change relative to the extraction period. Gaps can also occur during transitions periods for one coding system to another, e.g., during the transition period from ICD-9 to ICD-10, the hospital is not legally obliged to code the diagnosis codes for one calendar year. Software bugs occurring for a limited period of time can also create data gaps. Recording of some clinical observations may also be discontinued in recent data, for example, to ease the nurses' workload. \newline  RECOMMENDATION: Whenever known at extraction, such large impact hospital changes can be documented in the extraction documentation. This warns the researcher of the usefulness of extracted features for the prediction task. Exploration of trends over time in the recording of each feature (hospital-wide, but also per medical ward) detects these gaps in data. During data preparation, possible decisions are to not use unreliable features, proceed with missing data imputation or redefine the period of interest or the cohort of interest. \newline  METRIC: Measurement process - completeness; Documentation; WW: Completeness\\
\bottomrule
\end{longtable}
\endgroup{}

\begingroup\fontsize{7}{9}\selectfont

\begin{longtable}[t]{>{\raggedright\arraybackslash}p{3cm}>{\raggedright\arraybackslash}p{14cm}}
\caption{\label{tab:tbl-recommendations-feat-not-ts-2}FEATURE ENGINEERING Recommendations (not time-sensitive) Related to Feature Availability and Aggregation. ICD: International Classification of Diseases; CCS: Clinical Classifications Software; CCSR: Clinical Classifications Software Refined; METRIC = METRIC cluster and dimension mapping; WW = Weiskopf and Weng dimension mapping}\\
\toprule
Problem & Description and recommendation\\
\midrule
Ward-specific data recording patterns & DESCRIPTION: Difference between wards will exist in compliance of registration of specific care items and adherence to registration guidelines. These differences may be due to staffing requirements, or may be due to perceived clinical relevance for their specific patient population. Specific scores and scales may be applicable only to specific patient groups. Missingness patterns of extracted items may therefore differ between wards. \newline  RECOMMENDATION: During data preparation, missing data patterns can be explored in each ward and appropriate decisions can be taken depending on context: either impute missing data or engineer features in a contextually relevant manner, e.g. using GCS scores only in appropriate patient groups. \newline  Measurement process - completeness; WW: Completeness\\
Weekend effects result in missing data & DESCRIPTION: Due to staffing constraints, certain clinical items (e.g., registered by nurses) can be recorded with a lower frequency during weekends or public holidays. \newline  RECOMMENDATION: Exploring the extracted items or features trends over days of the week can help with detecting such trends. During data preparation, creation of separate features for the days of the week (Monday to Sunday) or weekend flags, eventually together with interactions with the features that display weekend gaps, can help the model distinguish between important patterns during weekdays vs. weekend days. \newline  Measurement process - completeness; WW: Completeness\\
Feature set definition is too restrictive & DESCRIPTION: Depending on the project approach, researchers might have to define the set of clinical items to extract at the start of the project. Typically, the important predictors for the outcome of interest are selected, based on literature review and/or input from clinical experts. It can though prove interesting to model the prediction task using a competing events framework \citep{gao2024comparison, albu2024comparison}. Auxiliary outcomes are often used as a form of regularization in deep learning applications \citep{tomavsev2021use}. In these situations, extracting items with predictive value for the competing events or auxiliary tasks can prove useful. Additionally, missing data imputation of extracted features can benefit of extraction of more features that have value in imputing the features of interest in the prediction task. \newline  RECOMMENDATION: Whenever possible, extraction of a larger set of clinical items of interest can benefit the data preparation process. \newline  METRIC: Informativeness - feature importance; WW: Completeness\\
Overmapping during extraction can dilute the predictive signal & DESCRIPTION: Although terminology mapping during data extraction can reduce significantly the time spent during data preparation, mapping into too broad categories can dilute the predictive signal. For example, ICD to CCS or CCSR mapping might be performed at extraction. Depending on the outcome to be predicted, history of specific ICD codes might prove more predictive than broader categories. \newline  RECOMMENDATION: Both the more granular extraction as well as the broader mapping could be made available during extraction, e.g., both ICD and CCS codes. \newline  Informativeness - feature importance; WW: Concordance\\
Use of hospital aggregate features & DESCRIPTION: The prediction task might benefit of the use of aggregate ward-level features, such as bed occupancy or nurse/clinician workload at the prediction time, typically aggregated per calendar day. Such features can, in principle, be calculated during data preparation based on the extracted data. At the prediction time in clinical implementation though, the prediction API should receive all data for all patients hospitalized at prediction time for the calculation of patient census or bed occupancy. This constitutes an unnecessarily heavy data transfer for making predictions for one patient at one point in time. Moreover, if the inclusion/exclusion criteria are applied at extraction (e.g., patients on mechanical ventilation), the researcher does not have access to the precise number of patients that occupy beds in the ward, as not all patients are on mechanical ventilation. \newline  RECOMMENDATION: It is more convenient to foresee such aggregate features during data extraction, considering that the same logic used at extraction can be reused at clinical implementation prediction time. Reusing the same database queries at prediction time reduces the need for large data transfers and exposure of all hospitalized patient data to the prediction API. \newline  Informativeness - feature importance; Measurement process - completeness; WW: Completeness\\
\addlinespace
Use of patient aggregate features & DESCRIPTION: Baseline features capturing aggregates of previous patient history, as for example historical comorbidities often prove useful for the prediction task. These can be calculated either during data extraction or at prediction time. Calculating these during data preparation comes though with downsides. First, the extraction might be limited to a specific time range, for example admissions from 2014 to 2020. Patients admitted in 2014 will suffer of left censoring for historical comorbidities, as their admissions prior to 2014 are not included in the extracted data. Second, at clinical implementation time, data for all patient admissions from the past would have to be sent to the prediction API, which might result in unnecessarily large data transfers over the network, especially if predictions are renewed often for a large number of patients. Last, when the exclusion/inclusion criteria are defined during data extraction, not all prior admissions for a patient are available (a patient undergoing surgery during current admission might have had prior admissions without surgery). \newline  RECOMMENDATION: Preferably, these would be calculated during data extraction and identically reproduced in clinical implementation. Within certain limitations, these can also be derived at preparation time. \newline  Informativeness - feature importance; Measurement process - completeness; WW: Completeness\\
\bottomrule
\end{longtable}
\endgroup{}

\begingroup\fontsize{7}{9}\selectfont

\begin{longtable}[t]{>{\raggedright\arraybackslash}p{3cm}>{\raggedright\arraybackslash}p{14cm}}
\caption{\label{tab:tbl-recommendations-feat-ts-1}FEATURE ENGINEERING Recommendations (time-sensitive) Related to Timestamp Integrity. EHR: Electronic Health Record; ICD: International Classification of Diseases; AKI: Acute Kidney Injury; CLABSI: Central Line-Associated Bloodstream Infections; ICU: Intensive Care Unit; METRIC = METRIC cluster and dimension mapping; WW = Weiskopf and Weng dimension mapping}\\
\toprule
Problem & Description and recommendation\\
\midrule
Timestamps shifting for anonymization & DESCRIPTION: As part of the anonymization process applied at data extraction, date shifting can be applied to deidentify the date of patient admissions \citep{johnson2016mimic, johnson2023mimic, maletzky2022lifting}. This process should preserve the sequence of admissions of a patient, the sequence of events within an admission, as well as seasonal features (e.g., time of day, day of the week, month of year). However, the process can introduce artifacts and render specific feature extractions during data preparation impossible (e.g. bed occupancy, nurse workload). \newline  RECOMMENDATION: The anonymization process applied at extraction should be tested using unit/integration tests in the data extraction framework. Whenever this process renders extraction of certain features unfeasible at data preparation, this should be documented. \newline  METRIC: Timeliness - currency; Documentation; WW: Currency\\
Timestamps when data are available in the system for real-time prediction differ from the clinically relevant timestamp & DESCRIPTION: The clinically relevant timestamp reflects the time that best corresponds to the patient’s actual physiological state, and can differ significantly from the data recording timestamp. For laboratory (and microbiology) results \citep{sauer2022leveraging, tomavsev2021use}, the relevant timestamps in the clinical workflow are: order time (when the clinician orders a lab test), planned collection time, collection execution time (when the nurse collects the sample), preliminary result availability time and final (validated) result availability time. Typically, the sample collection time and final result time are extracted from the system. For example, the time when a positive microbiology result is available in the laboratory system is often more than one day after the blood collection by the nurse; the latter is the clinically relevant timestamp because the result reflects the patient state at blood collection time (one day earlier than the result availability). Using the wrong timestamp in prediction will create temporal leaks, but its impact on the individual predictions will differ depending on how features are aggregated in time windows and how often the predictions are renewed (e.g., predictions renewed hourly, every 24 hours or trigger based - when new data become available in the system). Ward transfers can also be entered in the system before or after the actual patient care starts, which represent the clinically relevant time, e.g., transfer to ICU differs from the start of patient monitoring \citep{de2024table}. \newline  RECOMMENDATION: Most of the times (and especially for lab results), the relevant timestamps can be extracted from the system. Whenever not possible, the extraction documentation should mention the meaning of the extracted timestamp and its usual recording procedure and possible delays in recording. To avoid time leaks at implementation time, the registration time in the system (result time for lab results) should be used for features included in the prediction model. On the contrary, for outcomes calculated retrospectively based on extracted lab results (e.g., AKI, sepsis, CLABSI), the clinically relevant timestamp (sample collection time for lab results) should be used for outcome calculation, as mentioned in the Table 5. \newline  METRIC: Timeliness - currency; Documentation; WW: Currency\\
Extracted timestamps may reflect the time of the last modification rather than the original creation time of an item. & DESCRIPTION: Extracted timestamps may correspond to the creation time if the item has never been modified or to the last modification time if changes occurred. Most EHR systems do not maintain a full history of modifications, and even when they do, this information is rarely extracted. For example, lab results go through multiple updates: preliminary result, updated preliminary result, validation by a technologist, and final validation by a biologist. While the result value rarely changes, modifications are possible, and updates typically occur within minutes or hours. In contrast, demographic data, such as a patient’s home address, may change over months or years. Since only the most recent address and its last modification timestamp are usually extracted, historical changes might be missing. If spatial exposures are relevant to a prediction task, and extracted data spans over multiple years in which changes likely happened, important information might be missing. \newline  RECOMMENDATION: Extraction documentation should clearly reflect the meaning of extracted timestamps, e.g., creation / modification / result validation time. Insights from system users on how frequently modifications occur can help researchers determine whether the extracted data accurately reflects historical records. \newline  METRIC: Timeliness - currency; Documentation; WW: Currency\\
No date or timestamps available & DESCRIPTION: Some items are not extracted with a timestamp or have date but no time. For example, ICD codes are linked to admission and usually not timestamped; pharmacy medication orders have a date associated, but no timestamp. \newline  RECOMMENDATION: A conservative approach is to encode the timestamp at the end of the relevant time unit (e.g., ICD codes at the end of the admission \citep{tomavsev2021use}; medication orders end of day) or a less conservative approach by assigning a typical time of registration (e.g., a very high percentage  of medications are ordered before noon, assign 12:00 PM). \newline  METRIC: Timeliness - currency; WW: Currency\\
\bottomrule
\end{longtable}
\endgroup{}

\begingroup\fontsize{7}{9}\selectfont

\begin{longtable}[t]{>{\raggedright\arraybackslash}p{3cm}>{\raggedright\arraybackslash}p{14cm}}
\caption{\label{tab:tbl-recommendations-feat-ts-2}FEATURE ENGINEERING Recommendations (time-sensitive) Related to Temporal Leaks and Timing Errors. EHR: Electronic Health Record; ETL: Extract, Transform, Load; CPT: Current Procedural Terminology; ICU: Intensive Care Unit; METRIC = METRIC cluster and dimension mapping; WW = Weiskopf and Weng dimension mapping}\\
\toprule
Problem & Description and recommendation\\
\midrule
The extracted timestamp does not reflect the time when data are recorded in the system. & DESCRIPTION: The precise timestamp when items are recorded (or even modified) in the system might not be extracted, or might not be stored in the EHR database for extraction. Typical timestamps differences occur for nurse and clinician observations and coded procedures. For example, nurses might record observations in the patient chart after they finish a shift and they might record an approximation of the time they observed the patient, e.g., at 10:13 the nurse inserts a wound care observation with the time of observation: 8:20 (an estimation of the time the nurse visited the patient). Coded procedures (CPT codes), such as catheter insertion, can be recorded at or after the time of the procedure’s completion and the timestamp assigned to the procedure is the execution time or a close approximation of it. However, CPT codes can also be reviewed and documented retrospectively, even after patient discharge, when coding is reviewed for billing, based on patient's documentation throughout the hospitalization. There can be variation in the recording process, even within the same procedure code: sometimes procedures codes are recorded at the time of execution and other times after patient discharge. The discrepancy between the data available at model development and data available at prediction time in the clinical implementation may be larger for coded procedures compared to the nurse tasks, because of their review for billing purposes. \newline  RECOMMENDATION: The extraction documentation should mention the meaning of the extracted timestamp and its usual recording procedure and possible delays in recording. These are though difficult to document exhaustively, as the nurse flow might differ in different wards and coding procedures might vary widely, situations in which understanding of the data recording procedures will require communication with the clinical experts. \newline  METRIC: Timeliness - currency; Documentation; WW: Currency\\
Exact-hour timestamps often do not reflect the time when data are recorded in the system & DESCRIPTION: If a large proportion of timestamps in a date/time field appear as exact hours (e.g., 08:00:00 or 23:00:00), it may indicate that the extracted timestamps deviate from the time of recording in the system. Nurses might complete the tasks while they are at patient's bedside (09:12:30), but they record them in the EHR software after they finish their round (at 11:31:30), defaulting the execution time to a pre-set planned time, such as the start of a nurse’s shift or a standard scheduled time (08:00:00), instead of the time the task was executed in the system. In this example, the extracted timestamp might be 08:00:00, while the time available in the system for prediction would be 11:31:30. Depending on the EHR software and the hospital flow, an exact-hour timestamp could also mean that the task was planned, but not executed (when task execution status is not available in the extraction). \newline  RECOMMENDATION: The data extraction documentation of the extracted timestamp columns should mention the exact meaning of the extracted timestamps: execution, planning or registration time, or a combination thereof. Although the procedures for data recording might differ from ward to ward, whenever possible, the hospital flow should be documented and understood. When the exact-hour timestamps do not represent the meaningful timestamps, but the clinical items recorded are of predictive value, encoding the timestamps at the end of the relevant time unit (e.g., end of nurse shift) for model training would help avoiding temporal leaks. \newline  METRIC: Timeliness - currency; Documentation; WW: Currency\\
Temporal leaks due to extraction errors & DESCRIPTION: Temporal leaks can also occur because of errors during the ETL process, for example introduced by incorrect time zone handling, incorrect joins or time based filtering clauses. These are typically difficult to spot during data preparation. If the temporal leaks have a profound effect on the prediction performance, features that are not expected to have very high importance might appear as most important (e.g., history of bacteremia being the most important feature in predicting bacteremia during current admission might point to incorrectly leaking the current admission in the extraction of the historical feature). Some measurement timestamps may fall outside the admission period, occurring either before admission or after discharge. \citep{miao2023data}. While a small number of such cases can be normal in the workflow (registrations done on preadmission or discharge decision prior to lab result being returned by the laboratory), a large number of such cases might point to date shifts during extraction. Generally temporal leaks will impact the model performance, with the impact varying depending on how much time has been leaked. \newline  RECOMMENDATION: The primary way to safeguard from errors is rigorous testing of the ETL process. Manually checking a random sample of the admissions for the sequentially  of events \citep{tomavsev2021use} (e.g., items extracted from ICU system being present before ICU admission) can detect problems not detected during ETL testing. Building a tentative model on the training data and inspecting feature importance can reveal high impact temporal leaks. \newline  METRIC: Timeliness - currency; WW: Currency\\
Temporal leaks due to incorrect date/time handling or incorrect linking during data preparation. & DESCRIPTION: Incorrect time zone handling, incorrect joins, time based filtering or grouping and ordering items temporally can introduce temporal leaks during data preparation. Such leaks can distort predictions by providing information that would not be available at the time of prediction in clinical implementation. For example, misalignment of time zones may shift timestamps incorrectly, making future events appear as if they occurred earlier. Incorrect joins between tables (e.g., linking lab results to patient records without proper date/time filtering) can introduce data that should not be available at prediction time. \newline  RECOMMENDATION: Thorough unit testing of the data preparation pipeline is essential to prevent temporal leaks. For example, when predictions are generated at specific triggers (every 6 or 24 hours), testing should ensure that processing a full admission retrospectively in the data pipeline produces the same results as processing data incrementally up to each trigger time. This step helps confirm that only information available at a given prediction time is used, preserving the model’s real-world applicability. \newline  METRIC: Timeliness - currency; WW: Currency\\
\bottomrule
\end{longtable}
\endgroup{}

\begingroup\fontsize{7}{9}\selectfont

\begin{longtable}[t]{>{\raggedright\arraybackslash}p{3cm}>{\raggedright\arraybackslash}p{14cm}}
\caption{\label{tab:tbl-recommendations-clean-1}DATA CLEANING Recommendations Related to Outliers and Encoding Issues. EHR: Electronic Health Record, ETL: Extract, Transform, Load; ICU: Intensive Care Unit; CPR: cardiopulmonary resuscitation; METRIC = METRIC cluster and dimension mapping; WW = Weiskopf and Weng dimension mapping}\\
\toprule
Problem & Description and recommendation\\
\midrule
Non-numeric values in a numeric field & DESCRIPTION: Non-numeric values in a numeric field can encode information about the approximate numerical value. Discarding these (or implicitly transforming such fields to numeric and automatically creating missing values) would result in loss of information. A typical example is laboratory values that are capped at a specific threshold due to the detection limit of the assay, meaning measurements beyond this value are not precisely quantified \citep{albert2008modeling}. For instance, D-dimer results may be reported as '> 7650' when they exceed the assay’s measurable range. \newline  RECOMMENDATION: The capped value should be extracted as is and preserved as numeric during data preparation, as it contains more information than a missing value. \newline  METRIC: Measurement process - device error - accuracy; WW: Correctness\\
Values outside a physiologically plausible range (outliers) & DESCRIPTION: Manually entered numerical values can contain errors. The amount and magnitude of errors depends on the field definition in the EHR software, e.g. some software might not allow values of body temperature higher than 45 °C or dangerous drug doses, displaying an error or warning on the screen when recording such values, while others will allow any numerical value. \newline  RECOMMENDATION: Exploration of feature ranges and outliers should be done during data preparation. For all numerical values (e.g., laboratory results, vital signs), a physiologically plausible range can be defined in the data preparation pipeline, and such values can be deleted, or replaced with missing values to be imputed at a later stage in the pipeline. This procedure is called outlier capping \citep{tomavsev2021use}. We do not recommend manually correcting such errors, or applying complex rules that cannot be programmatically reproduced at clinical implementation. \newline  METRIC: Measurement process - device error - accuracy; Measurement process - human-introduced error - outliers; Consistency - logical consistency - plausibility; WW: Correctness; Plausibility\\
Implausible or inconsistent values within a time series & DESCRIPTION: Numeric measurements within a patient over time constitute time series and outliers can be explored in the context of previous or future values of the time series. For example, slipped sensors used in ICU \citep{maletzky2022lifting}, as for example a slipped bladder catheter, can record body temperatures closer to room temperature for a period of time, e.g., 32 °C. While such values are physiologically valid for patients on targeted temperature management, they can be implausible in a specific patient context if they occur only for some minutes or hours. Another example constitutes a heart rate of 0 used to mark admissions not linked to a real patient \citep{de2024table}, while in the ICU a heart rate of 0, within the context of previous normal heart rate followed by a normal heart rate can indicate a CPR procedure was done. Race, sex and age inconsistencies within patient but across several admissions can occur, e.g., for one admission the race is Hispanic, for another Caucasian \citep{miao2023data}. Presumably, these should be very rare. \newline  RECOMMENDATION: Corrections within a time series should be done during data preparation in a programmatic way (automated, reproducible, re-executable at prediction time), using only previous values from the same time series or from other features. For the temperature use case, a pragmatic approach can be to use only maximum values of temperatures within a time window and use a large enough time window, like 24 hours, within which it is unlikely to have only low temperature values except if the patient is on targeted temperature management. If inconsistencies are rare and a programmatic way of correcting these proves unfeasible (e.g., is the patient truly Caucasian or truly Hispanic?), a decision leave as is might be the best option. \newline  METRIC: Measurement process - device error - accuracy; Measurement process - human-introduced error - outliers; Consistency - logical consistency - plausibility; WW: Correctness; Plausibility\\
Character encoding problems & DESCRIPTION: Character encoding problems, e.g., MeniÃ¨re’s disease, can either be recorded as such in the EHR fields, due to a bug in the EHR software (that could have manifested only for a limited period of time) or can constitute a bug in the ETL process. \newline  RECOMMENDATION: Character encoding consistency in the ETL process should be ensured by unit/integration tests. Whenever the problem manifested in the EHR itself, correct mapping to the correctly encoded terminology should be foreseen in the data extraction or preparation. \newline  METRIC: Measurement process - human-introduced error - carelessness; Consistency - rule-based consistency - syntactic consistency; WW: Correctness\\
Inconsistent or incorrectly specified units & DESCRIPTION: Units for height, weight, lab results, and other numerical data \citep{miao2023data, ferrao2016preprocessing} can differ for different measurements, e.g., height in cm or in m in function of patient's age, or over time, e.g., blood glucose in mg/dL and mmol/L. In some EHR systems, units might be specified but might not correctly reflect the units used historically in past years of the data (if the relational database does not store a history of the linked units). \newline  RECOMMENDATION: Whenever units are correctly specified for all measurements, unit conversion can easily be applied during data preparation. Historical units inconsistencies can be detected by exploration of statistics over time (e.g., minimum, maximum, mean values) and corrected only for historical data (assuming the units recorded now will be maintained in future data, at implementation). \newline  METRIC: Measurement process - source credibility - traceability; Consistency - rule-based consistency - syntactic consistency; WW: Correctness\\
\bottomrule
\end{longtable}
\endgroup{}

\begingroup\fontsize{7}{9}\selectfont

\begin{longtable}[t]{>{\raggedright\arraybackslash}p{3cm}>{\raggedright\arraybackslash}p{14cm}}
\caption{\label{tab:tbl-recommendations-clean-2}DATA CLEANING Recommendations Related to Missing Data and Logical Inconsistencies. EHR: Electronic Health Record; ETL: Extract, Transform, Load; ICU: Intensive Care Unit; METRIC = METRIC cluster and dimension mapping; WW = Weiskopf and Weng dimension mapping}\\
\toprule
Problem & Description and recommendation\\
\midrule
Clinically implausible combinations of patient characteristics & DESCRIPTION: Even when values are within a plausible range for the entire cohort, they can be implausible for the specific patient. Examples include: length of 1m for a neonate; weight of 200g for an adult \citep{schwabe2024metric}, patient age at admission is 50 years but admission type 'new-born' \citep{miao2023data}, paediatric conditions for adults \citep{blacketer2021increasing}. Such examples will occur mostly in manually entered data. \newline  RECOMMENDATION: Exploration of feature ranges should be performed also within relevant subgroups for which the plausible ranges differ significantly. As EHR data extraction include a rather large sample size and these errors are rare, these can be set to missing and imputed at a later stage, provided that the imputation model uses relevant features in imputation (e.g., imputing length using weight and age); mean or median imputation (over all population) might prove as error prone as the initial error. Manual correction of such errors case-by-case would not provide a consistent correction strategy at model implementation time. Correction schemes that can be programmatically reproduced (like imputation) are preferred. \newline  METRIC: Measurement process - human-introduced error - carelessness; Consistency - logical consistency - plausibility; WW: Correctness; Plausibility\\
Syntactic variability for missing values & DESCRIPTION: Unknown values at the time of registration can be either left as an empty field, or entered as a free text with various values, such as 'Not Mapped', 'Unknown', 'Not specified', 'N/A', 'NA', '?', etc. \citep{miao2023data, ferrao2016preprocessing}. \newline  RECOMMENDATION: The values entered in a field should be explored and all values that represent an unknown value at the time of recording should be encoded as missing. \newline  METRIC: Measurement process - human-introduced error - noisy labels; Consistency - rule-based consistency - syntactic consistency; WW: Correctness\\
Missing values that can be fully recovered from other fields & DESCRIPTION: Some extracted data fields might encode, for specific values of the fields, similar or identical information. Admission ward, admission source and admission type are such examples. Missing values in one of these can, in some cases, be recovered from the other extracted fields. Miao et al \citep{miao2023data} encountered a situation of trivial imputation scenario, in which admission source was emergency room but admission type was missing. In this case, the admission type is also emergency. \newline  RECOMMENDATION: Whenever the imputation is trivial and easy to reproduce programmatically, this can be performed during data preparation. At the same time, a powerful missing data imputation algorithm, that effectively utilizes patterns in other features, might be able to detect and correctly impute such values, given enough correct examples exist in the extracted data. \newline  METRIC: Measurement process - completeness; WW: Completeness\\
Implausible missingness rates & DESCRIPTION: While EHR data will inherently exhibit missing values, missingness can be also an artifact of errors during data extraction or preparation. For dynamic prediction models, missing data patterns can be explored using an aggregation unit that is meaningful for communication with clinical experts, such as a patient day, even if the final model uses a different aggregation time window. This is typically done after feature engineering, and can, in certain situations, reveal feature engineering problems. For example, a missingness rate of around 30\% for patient age (which is known in EHR for almost all patients) can have as root cause a problem during linking identifiers, or even in the ETL anonymization process. No missing data in features in which missing data are expected can be due to an unintentional imputation during the ETL process. Missing data can also occur because of incorrect mappings during extraction. For example, if missingness rates in vital signs in ICU, where these are measured at highest frequency, are higher than in other wards, this could prove to be due to incorrect (or omitted) mapping during data extraction or preparation. \newline  RECOMMENDATION: Whenever the missingness rate is not in line with EHR capabilities or with the expected recording frequency in different wards or patient subgroups, the extraction and preparation logic of those features should be verified for errors. \newline  METRIC: Measurement process - completeness; WW: Completeness\\
\bottomrule
\end{longtable}
\endgroup{}

\section{Discussion and Conclusion}\label{discussion-and-conclusion}

In this work, we highlighted challenges and recommendations for the extraction and preparation of EHR data for predictive modeling. Adhering to the ``garbage in, garbage out'' principle, prediction models rely heavily on the quality and relevance of the input data to generate meaningful and reliable predictions. High-quality data extraction and preparation processes can support prediction models with clinical utility. While it is often argued that EHR ``data are not collected for research purposes'' \citep{van1991use, goldstein2020five, sauer2022leveraging, maletzky2022lifting, thuraisingam2021assessing, arbet2021lessons}, the fact that these data are sufficient for supporting the patient care suggests they could be adequate for developing prediction models. Even though the list of challenges may seem extensive, it should not discourage researchers new to working with EHR data. Not all challenges will apply to every project, and as EHR systems continue to evolve, many issues might affect only historical data and not current data at implementation time. Successfully leveraging EHR data requires both an understanding of its limitations and an appreciation of its potential. Despite its complexities, EHR data offer a rich, comprehensive source of real-world clinical information that can drive impactful research and improve patient outcomes. Applied prediction models can exploit the comprehensive clinical information that is not always readily available to all healthcare professionals, such as nurses, doctors, hygienists, and other therapists, and have proven practical applicability \citep{leeclinical}.

Data extraction and preparation for predictive modelling using EHR data are resource-intensive processes, with time and cost varying depending on the maturity of the extraction framework, the prediction task and the team's experience in working with EHR data. It is estimated that this phase (which generally includes collaboration with data extraction engineers and clinical experts), takes up to three months for an entire team \citep{tomavsev2021use}, but such estimations will be highly dependent on the maturity of the extraction process and of the team's experience with EHR data. Data quality often varies significantly across EHR systems and extraction processes, as noted by Weiskopf et al. \citep{weiskopf2013methods}. Issues such as data gaps, temporal leaks, incorrect linking, inconsistencies in clinical concept and terminology mapping can affect the quality of extracted datasets and compromise the model's performance and applicability. To address such challenges, we proposed a list of practical recommendations informed by our experiences with EHR data and insights from published studies. We organized the challenges and recommendations into: cohort definition, outcome definition, feature engineering and data cleaning; the first three categories can also be consulted when planning a project. A clear definition of the prediction task or research question, of the intended use and the intended users of a prediction model are a first critical step for defining the outcome, the cohort and the features of interest \citep{arbet2021lessons, honeyford2022challenges}. It is not uncommon to deem the EHR data inadequate for the prediction task, before proceeding with the model building phase \citep{thuraisingam2021assessing}.

For cohort definition, we recommend extracting a broader patient context beyond the immediate focus, assessing the completeness of data used for inclusion/exclusion criteria and its availability at prediction time, preventing omissions or duplications, and careful definition of episodes of interest for prediction. Outcomes can be derived in different ways, each with advantages and shortcomings. We generally advise against the use of ICD codes (as these can be up- or undercoded and are generally not timestamped), unless carefully assessed as appropriate for the prediction task. Good understanding of hospital's processes, thorough verification of outcomes derived in code, manual inspection of labels and agreement between data sources can also prevent incorrect outcome definition. Mapping terminology to coding systems (e.g., ICD, CPT, LOINC, SNOMED-CT or others) can facilitate feature engineering. However, not all items in the EHR system are aligned with standardized terminologies (e.g., LOINC codes for lab results are often not used). Implementing terminology mapping during the data extraction phase (e.g., through OMOP CDM) can significantly reduce the effort required during data preparation. High-quality data extraction documentation and good understanding of the underlying clinical concepts and healthcare processes are essential to support feature engineering. Data exploration can reveal unaddressed problems and further inform the construction of meaningful features. For feature engineering in prediction settings, the timestamp when a clinical item is available in the system is of greatest interest, as this is the time at which predictor values become available for prediction in clinical practice. Good documentation and correct interpretation the extracted timestamps will prevent temporal leaks. Thorough verification of the extraction and preparation processes (using manual and/or automated tests), data exploration and reproducible data cleaning can safeguard against data quality issues.

Clinical assessment of the relevance and the sequence of extracted features and outcome within a patient admission by manual verification of a random sample of admissions can further help detecting problems \citep{tomavsev2021use}. The solutions to specific issues can be implemented at either the extraction or preparation stage. Applicable to both data extraction and preparation are good understanding of the underlying data structure and current and historical healthcare processes, collaboration with relevant experts in conducting the work \citep{tomavsev2021use, honeyford2022challenges}, and ensuring a qualitative process of extraction and preparation, supported by unit tests. We recommend maintaining consistency in data extraction and preparation between model training and clinical implementation. Exceptions might though exist for correcting historical data problems that are not expected to reoccur in future data.

We hold the opinion that, in the context of prediction models, the extraction process should not attempt to correct errors residing in the EHR database in the attempt to align the data to the clinical reality, but it should reflect the information from the EHR, presented in a simplified format. We recommend that data cleaning steps are performed during data preparation, in a pragmatic and programmatic manner that can be reproduced at implementation time. Extracting and preparing the training data in a different manner than for clinical implementation (e.g., temporal leaks) poses the risk of potential overoptimistic evaluation, in the light of which the model's performance when implemented in clinical practice will be lower than expected. We acknowledge that there are divergent views on this topic and that in the context of inferential studies, when the analysis does not need to be reproduced on future data, corrections during data extraction might be preferred. At the same time, multi-purpose extractions (for both prediction and inferential studies) pose the challenge of solving this divergent view.

While the level of detail for reporting data extraction and preparation in published prediction studies varies, we do not provide specific recommendations on this aspect. Some researchers advocate for comprehensive documentation of these processes \citep{de2024table, chin2023guiding, hernandez2020minimar}, others emphasize the importance of sharing the data preparation code to ensure transparency \citep{goldstein2024improved, johnson2017reproducibility}. Data extraction for public datasets is generally documented in a separate publication. This can be complemented by sharing the data preparation code and describing in the main article the key differences between the raw extracted data and the final prepared dataset \citep{de2024table}. Each strategy has its advantages, depending on the audience, journal requirements, and the desired tradeoff between transparency and conciseness. While we advocate for transparent reporting, this article does not specifically address reporting guidelines for data extraction and preparation.

Our work has several limitations. First, it is based on our experiences and a selective literature review that cannot be exhaustive. We acknowledge that every project will face use case specific challenges. Insights from other research groups working with EHR data would likely highlight additional challenges and recommendations that we may have overlooked, offering a more complete set of recommendations. Second, our focus was on single-site structured EHR data. Extensions to multi-center datasets or unstructured data are possible. Multi-center datasets pose additional challenges with regards to aligning clinical concepts and following the same patient in different hospitals and/or general practitioner systems. Patient care happens across multiple systems, and single-site extractions provide only a fragmented view of the entire patient care. Third, we provide recommendations for problems that can impede the model implementation in practice, but we do not cover explicitly post-implementation challenges, although some of our recommendations can inform on monitoring checks that can be implemented, which remains a subject for medical device post-market surveillance. Fourth, while we emphasize the importance of high-quality data extraction and preparation as the foundation of reliable predictive models, we do not assess the impact of each challenge on the final prediction model. The impact will likely depend on the magnitude of the problem, the subgroups in which it manifests, the prediction task or even the type of model (e.g., tree-based models are generally resilient to outliers). While the impact of dataset size, missing data and outcome definition on the model performance have been studied \citep{van2022developing}, the impact of other steps in the data preparation procedure remains unknown. Researchers identify and resolve problems during data preparation without assessing their impact on model performance, which we acknowledge having done the same. Research on measurement error \citep{luijken2020changing} demonstrates that inconsistencies in predictor definitions between training and test data can affect model calibration. A similar impact may occur if correction strategies differ between model training and clinical implementation. Finally, although we focused on models with clinical applicability, by recommending an implementation-ready process to avoid discrepancies between the training data and future data due to data extraction and preparation, we did not specifically address model generalization to new hospital settings. The requirements for reproducing the data extraction and preparation can vary significantly based on the EHR software in use (whether from the same or different vendor), the data extraction platform, and the hospital workflow and data registration procedures. Sendak et al \citep{sendak2017barriers} estimate ``approximately 75\% of the effort invested in the initial data preparation for developing prediction models must be reinvested for each hospital''. We argue that the estimation would vary widely, with standardized extractions from the same EHR software potentially requiring less time. Further research could focus on extending the list of challenges and recommendations based on the experience with EHR data of other research groups, extending the scope to larger extraction contexts or assessing the impact of erroneous or suboptimal extraction and preparation on the final model.

The extensive list of challenges and practical recommendations for EHR data extraction and preparation presented here is intended for improving the quality of research and the practical applicability of clinical predictive models. Since all modeling efforts begin with the underlying data, failing to address data quality issues risks producing unreliable and non-generalizable models. Our focus extends beyond the initial data curation stage in the AI life cycle \citep{ng2022ai}; we also address early-stage issues that can ultimately negatively impact model deployment. Recognizing that there is no ``one-size-fits-all'' solution, our list of challenges and recommendations, though not exhaustive, is comprehensive enough to support many EHR-based prediction projects. Implementing the strategies applicable to each project can ultimately enhance robustness, reproducibility, and real-world impact of EHR-based prediction models.

\section{Abbreviations}\label{abbreviations}

AKI: Acute Kidney Injury

API: Application Programming Interface

AVPU: Alert, Voice, Pain, Unresponsive

CAM: Confusion Assessment Method

CCS: Clinical Classifications Software

CCSR: Clinical Classifications Software Refined

CDC: Centers for Disease Control and Prevention

CLABSI: Central Line-Associated Bloodstream Infection

COVID-19: Coronavirus Disease 2019

CPT: Current Procedural Terminology

DRS: Delirium Rating Scale

ETL: Extract, Transform, Load

EHR: Electronic Health Record

FiO2: Fraction of Inspired Oxygen

HIS: Hospital Information System

ICD: International Classification of Diseases

ICU: Intensive Care Unit

NA: Not Applicable

OMOP CDM: Observational Medical Outcomes Partnership Common Data Model

PEEP: Positive End-Expiratory Pressure

VAE: Ventilator-Associated Event

\bibliographystyle{plainnat}
\bibliography{references}

\end{document}